\documentclass[runningheads]{llncs}
\usepackage[T1]{fontenc}
\usepackage{graphicx}
\usepackage{booktabs}
\usepackage{multirow}
\usepackage{xcolor}
\usepackage{hyperref}
\newcommand{\orcidlink}[1]{\href{https://orcid.org/#1}{\includegraphics[height=9pt]{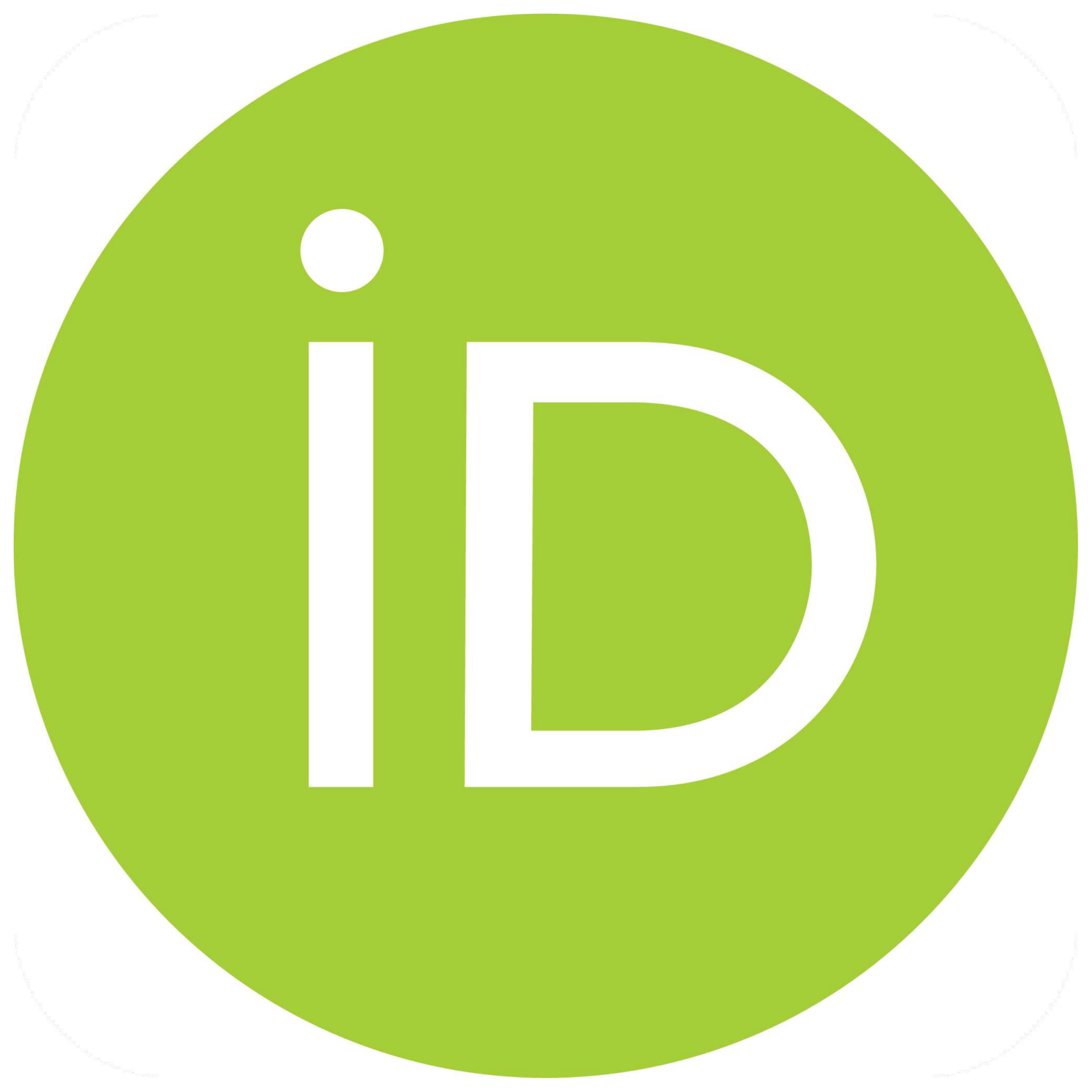}}}

\begin{document}
\title{From Recall to Reasoning: Automated Question Generation for Deeper Math Learning through Large Language Models}

%
%

\author{Yongan Yu\orcidlink{0009-0006-3264-1170} \and Alexandre Krantz\orcidlink{0009-0006-5772-5346} \and Nikki G. Lobczowski\orcidlink{0000-0002-9018-2957}}

\institute{McGill University, Quebec, Canada \\
\email{\{yongan.yu, alexandre.krantz\}@mail.mcgill.ca}, \email{nikki.lobczowski@mcgill.ca}}

\titlerunning{Question Generation for Deeper Math Learning via LLMs}
\authorrunning{Y. Yu et al.}
\maketitle              

\begin{abstract}
 Educators have started to turn to Generative AI (GenAI) to help create new course content, but little is known about how they should do so. In this project, we investigated the first steps for optimizing content creation for advanced math. In particular, we looked at the ability of GenAI to produce high-quality practice problems that are relevant to the course content. We conducted two studies to: (1) explore the capabilities of current versions of publicly available GenAI and (2) develop an improved framework to address the limitations we found. Our results showed that GenAI can create math problems at various levels of quality with minimal support, but that providing examples and relevant content results in better quality outputs. This research can help educators decide the ideal way to adopt GenAI in their workflows, to create more effective educational experiences for students. 

\end{abstract}

\section{Introduction}
With the rapid advancements in LLMs, AI has significantly impacted educational development \cite{osti_10440200,antu2023using}, transforming how academic content is created and delivered. Recent studies demonstrate LLMs can enhance learning experiences across various scenarios \cite{denny2024computing,li2024bringing}. Particularly, recent studies have explored AI applications, including personalized suggestions \cite{xiong2024review}, and learning behavior analysis \cite{batool2023educational}.

While over 60\% of educators have experimented with ChatGPT, less than 20\% feel adequately prepared to integrate it effectively \cite{Chatgptreview}. One application of LLMs that could significantly support teachers is Question Generation (QG). A high-quality LLM QG tool could significantly reduce the workload of educators \cite{de2020towards}, as it would free up the time spent on creating problem sets and answers. It could also result in more practice problems for students, enhancing their learning experience. From a technical standpoint, QG  is an existing sub-field of natural language processing (NLP), focused on enabling the automated creation of educational content directly from reference material, such as textbooks \cite{elkins2024teachers}.

Despite the transformative potential of AI in education, its integration remains underutilized, especially in automated question generation. A need-finding study by Wang et al. (2023) \cite{WANG2024102275} reveals that educators often express reservations about adopting AI tools, citing concerns about the relevance and quality of AI-generated content. Moreover, existing automatic QG tools are not widely used in classrooms due to their limited range in types and difficulty levels \cite{kurdi2020systematic}. Most systems primarily produce simple recall questions, failing to sufficiently challenge students or promote deeper cognitive processing.

\textbf{Problem Statement} Our study investigates the gap between AI capabilities and effective educational implementation by answering: \textbf{RQ1}: \textit{What impact does increased contextual information have on the quality and cognitive depth of LLM-generated questions?} and \textbf{RQ2}: \textit{How can LLM-based question generation systems be designed to produce questions across varying levels of cognitive depth? }We hypothesize that developing a context-aware AI framework integrating established educational taxonomies \cite{irvine2021taxonomies} will enable the generation of high-quality, diverse, and cognitively appropriate questions that align closely with specific educational objectives and content. Our goal is to bridge the gap between AI capabilities and educational needs, potentially increasing educator confidence and the adoption of AI tools in classrooms for improved educational outcomes.

\section{Related-Work}

The rapid rise of GenAI, exemplified by ChatGPT's release in late 2022, has significantly impacted the educational landscape \cite{lo2023impact}, with the potential to automate 20-40\% of teachers' administrative tasks \cite{denny2024computing}. However, current GenAI models show limitations in math education. For instance, while effective for retrieving theorems, they struggle as conversational tutors and make errors even with elementary problems \cite{frieder2024mathematical}. These issues stem from a lack of logical reasoning \cite{mccarthy2022artificial}, highlighting the need for specialized AI systems in math education.

QG systems have evolved from simple recall questions \cite{wang2022towards} to more sophisticated models like "QG-Net" \cite{wang2018qg}, which uses recurrent neural networks to generate quiz questions from educational content. However, math-specific QG faces unique challenges, including limited benchmark datasets \cite{wu2023conic10k} that can lead to overfitting and question redundancy \cite{guo2024survey}. Our study aims to address these limitations by creating higher-quality math questions while building AI systems that can truly adapt to students' learning journeys.

\section{Study 1: Exploring Current Capabilities of GenAI}

We explored how GenAI could enhance educational practices, particularly in intelligent tutoring systems and automated quizzing \cite{shute199419}. Our goal was to integrate GenAI seamlessly into student and teacher workflows, ensuring reliability and context awareness. To achieve this, we investigated whether GenAI could generate comprehension questions relevant to course content and analyzed their cognitive depth using Bloom’s Taxonomy \cite{anderson2001taxonomy}.

\textbf{Study Design} We tested three "level of context" scenarios with progressively more information provided to the GenAI. Using a mathematical logic course covering topics such as satisfiability and the compactness theorem, we designed prompts instructing the model to generate five comprehension questions with answers. We employed \textsc{Gemini-1.5-Pro} \cite{reid2024gemini} with a temperature of 0 \cite{zhu2024hot} to minimize output variations. In Scenario 1 (minimal), the instructor provided only the course syllabus and a brief topic summary; in Scenario 2 (moderate), they added their notes for a specific class session; and in Scenario 3 (comprehensive), they included the syllabus, class notes, and references covering course material.

\textbf{Evaluation Metrics} A math student with expertise in logic evaluated outputs via three metrics: relevance (binary score indicating whether the question was within lesson scope), depth (score from 0-6 corresponding to Bloom's Taxonomy levels), and correctness (binary score assessing answer accuracy).

\begin{table}[t!]
\centering
\scriptsize
\caption{Performance comparisons for three context scenarios. Note: Values are presented as mean ± standard deviation.}
\renewcommand{\arraystretch}{1.2}
 \begin{tabular}{||c c c c||} 
 \hline
Context Scenario & Relevance & Depth & Correctness\\ [0.5ex] 
 \hline\hline
Minimal & \textbf{1.00 ± 0.00} & 2.60 ± 1.14 & 0.60 ± 0.55\\ 
Moderate & \textbf{1.00 ± 0.00} & 2.60 ± 1.82 & 0.80 ± 0.45\\
Comprehensive & 0.80 ± 0.45 & \textbf{2.40 ± 2.07} & \textbf{1.00 ± 0.00}\\
 \hline
 \end{tabular}
\label{table:Initial Results}
\end{table}

\textbf{Findings from Initial GenAI Testing} In this preliminary experimentation, our analysis revealed that GenAI indeed can create content that is relevant and high-quality with input support. Results from testing were surprising in the level of relevance because there was a declining trend in the relevance level of the output as we added additional context. However, we found the depth of the questions remained relatively constant, and the correctness of the answers generated improved as more context was provided, reaching perfect accuracy with comprehensive context.

\begin{figure*}[t]
\centering
\includegraphics[width=0.8\textwidth]{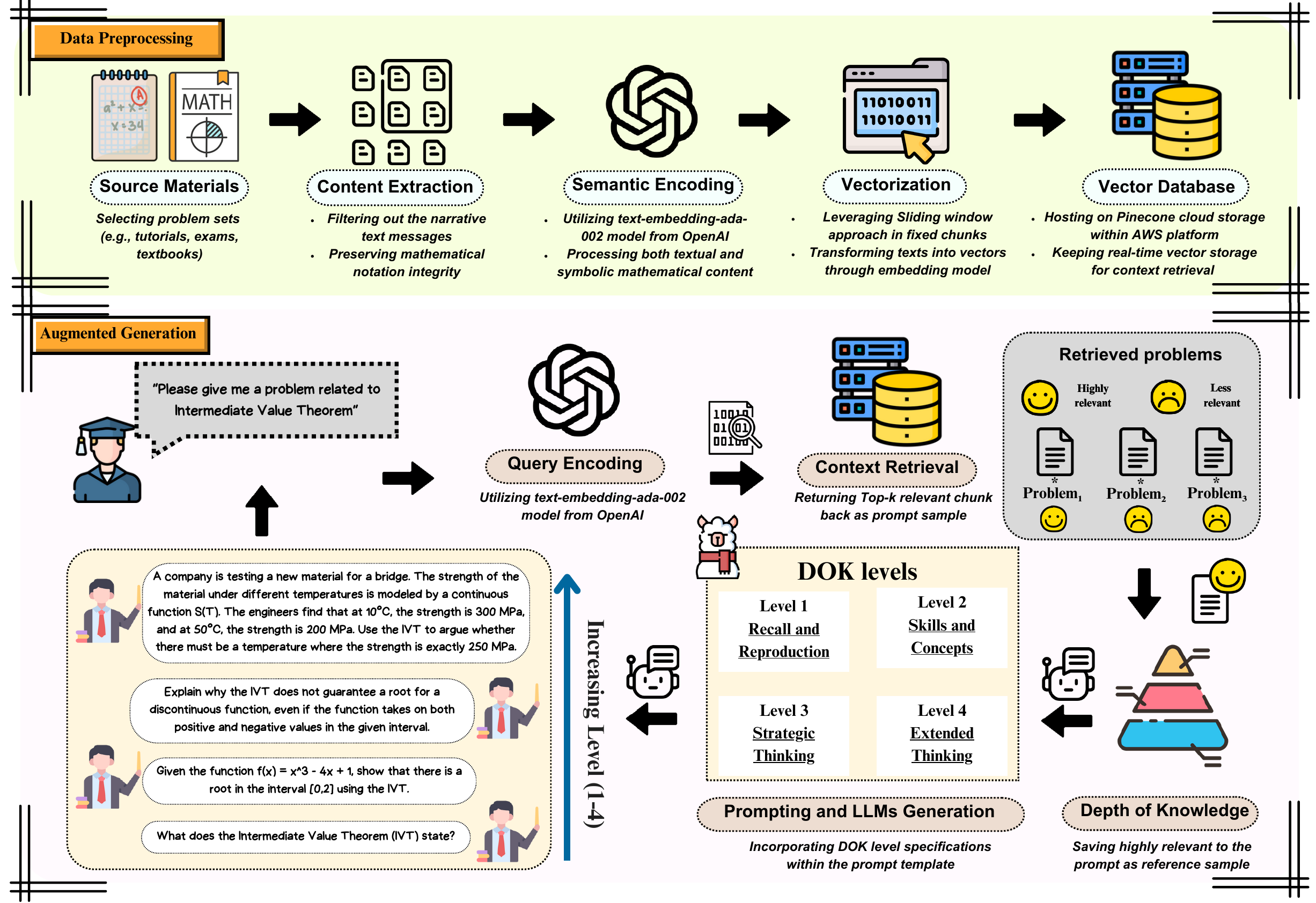}

\caption{Overview of the proposed framework with two core components}
\label{study2_framework}
\end{figure*}

\textbf{Advancing to the Next Stage: From Bloom's Taxonomy to DOK}
Our initial investigation revealed that additional context improved question correctness but not relevance, as the GenAI sometimes extrapolated and went beyond the bounds of what was taught in the course (i.e., creating "hallucinations"). Since instructors can generally verify answer correctness, we prioritized improving relevance in our next phase. We considered approaches like retrieval-augmented generation (RAG) \cite{lewis2020retrieval} to better ground outputs in provided materials, which could potentially resolve this issue by forcing the AI to source its generation from a "chunk" of the materials provided. Additionally, while Bloom provided a starting framework, we found Webb's depth of knowledge (DOK) \cite{webb2002depth} framework to be better suited for math education. DOK emphasizes task complexity and contextual knowledge application rather than just cognitive processes, aligning more effectively with mathematical problem-solving and curriculum standards. This framework enables a more precise mapping of question difficulty to the cognitive processes involved in mathematical reasoning, from basic recall to complex problem-solving.

\section{Study 2: Developing an Improved Framework}

Building upon prior findings, we introduced QG-DOK, a question generation framework integrating RAG with Webb's Depth of Knowledge to generate context-aware questions with varying cognitive depth levels, addressing limitations identified in our initial phase. Thus, our second research objective explored how GenAI can generate questions of varying difficulty, making them more adaptable to teachers in math education. Our system, illustrated in Figure \ref{study2_framework}, comprises two core components.

\begin{figure}[t]
\centering
\includegraphics[width=0.6\columnwidth]{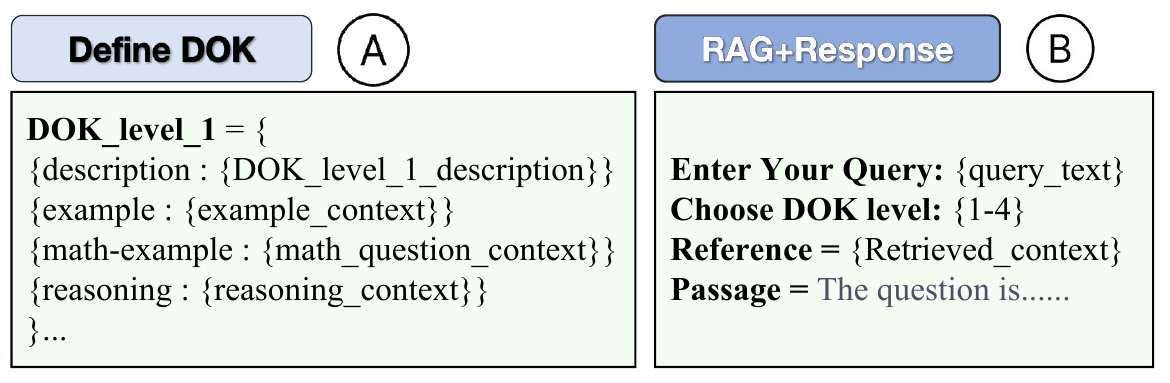} 

\caption{\textcircled{A} Level-1 prompt template example \textcircled{B} User input interface}
\label{figure:DOK Prompt}
\end{figure}

\textbf{RAG Framework }QG-RAG was implemented using a naive RAG framework \cite{gao2023retrieval} and integrated with DOK:
\begin{itemize}
    \item \textbf{Data preprocessing}: We gathered and refined mathematical content from textbooks, tutorials, and practice problems, then transformed it into vector representations through an embedding model. These vectors were stored in a database to facilitate efficient retrieval of semantically relevant materials.
    \item \textbf{Augmented generation}: When a user submits a query, the system retrieves relevant content from the vector database to provide contextual grounding. To adjust difficulty and cognitive depth, we incorporated four DOK levels into our question generation process, as shown in Figure \ref{figure:DOK Prompt} 
  \textcircled{A}: \textit{Recall and Reproduction (level-1)}: retrieving basic facts, definitions, and formulas with minimal cognitive effort; \textit{Skills and Concepts (level-2)} selecting appropriate methods and organizing information to solve routine problems; \textit{Strategic Thinking (level-3)}: reasoning, planning, and applying concepts in non-routine scenarios; \textit{Extended Thinking (level-4)}: making connections across concepts and solving complex, multi-step problems.
\end{itemize}

\textbf{Implementation Details} As shown in Figure \ref{study2_framework}, we embedded a corpus of mathematical content using the \textsc{text-bedding-ada-002} model. To enhance semantic relevance, documents were segmented into fixed-size chunks using a sliding-window approach \cite{wang2024searching}. UI functionality is shown in Figure \ref{figure:DOK Prompt} \textcircled{B}, users were prompted to input two key pieces of information: the mathematical concept they wish to explore and the desired DOK level. In second study, we evaluated three off-the-shelf LLMs for question generation with default temperature settings, including \textsc{GPT-4o} \cite{openai2024gpt4ocard}, \textsc{DeepSeek-V3} \cite{deepseekai2024deepseekv3technicalreport} and \textsc{Gemini-1.5-Pro} \cite{geminiteam2024gemini15unlockingmultimodal}.

\begin{table}[!t] 
    \centering
    \tiny
        \caption{Performance comparison of LLMs across three evaluation metrics in question generation. \textbf{Bold} and \underline{underline} indicate the highest and second-highest scores}
    \begin{tabular}{cl|cc|cc|cc|c}
        \toprule
        & & \multicolumn{2}{c|}{\textbf{Relevance}} & \multicolumn{2}{c|}{\textbf{DOK alignment}} & \multicolumn{2}{c|}{\textbf{Appropriateness}} & \textbf{PINC} \\
        \cmidrule{3-8}
        & & DOK & DOK+RAG & DOK & DOK+RAG & DOK & DOK+RAG & \\
        \midrule
        \multirow{4}{*}{\rotatebox[origin=c]{90}{\textsc{GPT-4o}}} 
        & Level 1 & \textbf{0.85} & 0.79 & \underline{0.81} & \textbf{0.78} & \underline{0.80} & 0.91 & \textbf{0.94}\\
        & Level 2 & 0.81 & \textbf{0.85} & 0.30 & 0.73 & 0.72 & 0.90 & \underline{0.93}\\
        & Level 3 & 0.79 & 0.81 & 0.52 & 0.71 & 0.74 & \textbf{0.95} & 0.92\\
        & Level 4 & 0.78 & 0.80 & 0.29 & 0.61 & 0.70 & 0.82 & 0.90\\
        \cmidrule{2-9}
        & Average & 0.81 & 0.81 & 0.48 & 0.71 & 0.74 & 0.90 & 0.92\\
        \midrule
        \multirow{4}{*}{\rotatebox[origin=c]{90}{\textsc{Deepseek-V3}}}
        & Level 1 & \underline{0.84} & 0.82 & \textbf{0.80} & 0.68 & \textbf{0.82} & \underline{0.89} & \textbf{0.94}\\
        & Level 2 & 0.80 & \underline{0.84} & 0.75 & 0.72 & 0.74 & 0.88 & 0.92\\
        & Level 3 & 0.77 & 0.80 & 0.63 & 0.78 & 0.72 & 0.85 & 0.90\\
        & Level 4 & 0.66 & 0.71 & 0.39 & 0.59 & 0.71 & 0.80 & 0.92\\
        \cmidrule{2-9}
        & Average & 0.77 & 0.79 & 0.64 & 0.69 & 0.75 & 0.86 & 0.92\\
        \midrule
        \multirow{4}{*}{\rotatebox[origin=c]{90}{\textsc{Gemini-1.5}}}
        & Level 1 & 0.80 & 0.75 & 0.75 & \underline{0.77} & 0.76 & 0.85 & 0.92\\
        & Level 2 & 0.75 & 0.78 & 0.49 & 0.68 & 0.68 & 0.82 & 0.90\\
        & Level 3 & 0.72 & 0.74 & 0.52 & 0.75 & 0.69 & 0.88 & 0.89\\
        & Level 4 & 0.63 & 0.73 & 0.46 & 0.58 & 0.65 & 0.75 & 0.92\\
        \cmidrule{2-9}
        & Average & 0.73 & 0.75 & 0.56 & 0.70 & 0.70 & 0.82 & 0.91\\
        \bottomrule
    \end{tabular}
    \label{study2-results}
\end{table}

\textbf{Evaluation Metrics} To assess our QG-DOK framework, we compared two implementations: (1) DOK, providing only DOK level definitions, and (2) DOK+ RAG, which retrieved relevant examples from a vector database. We evaluated using G-Eval \cite{liu2023g} to measure relevance, DOK alignment, and appropriateness. Additionally, we incorporated the Paraphrase n-gram Change (PINC) score \cite{scaria2024automated} to quantify lexical diversity.

\textbf{Findings from Improved Work} Our evaluation demonstrated that DOK+ RAG consistently outperformed DOK-only across all tested LLMs , particularly for higher-order thinking skills (DOK Levels 3 \& 4). DOK+RAG improved both relevance and appropriateness scores, with \textsc{GPT-4o} showing the most significant gains in appropriateness. Although DOK alignment showed mixed results, DOK+RAG generally improved depth accuracy at higher cognitive levels. Also, high PINC scores (average 0.92) indicated strong lexical diversity in question rephrasing.
Despite these improvements, challenges remained. The depth alignment at Level 2 was inconsistent across the models, suggesting that LLMs struggle with categorizing mid-level cognitive complexity. We also identified persistent issues in mathematical notation handling that would benefit from LaTeX-based representation in future implementations.

\section{Conclusion and Discussion}

Through two interconnected studies, our research provides insights into GenAI's potential in math education. \textit{Study 1} revealed that while GenAI can produce relevant questions using Bloom's Taxonomy, it struggles with higher cognitive levels and tends to generate plausible yet incorrect information when given more context.
Building on these findings, \textit{Study 2} introduced the QG-DOK framework, integrating Webb's DOK levels with RAG. By leveraging resources that educators are already familiar with, both quality and depth of generated questions improved. Our findings support earlier research suggesting that AI can effectively generate educational content but requires careful design to ensure cognitive depth and relevance \cite{denny2024computing}. The improvements in \textit{Study 2} address limitations identified in \textit{Study 1}, particularly in generating deeper thinking questions.

\section{Limitations}

Although our results highlight the potential of GenAI in educational content creation, a few limitations remain. Bloom's taxonomy, though often used to categorize cognitive understanding, has been critiqued for oversimplifying the interconnected nature of learning \cite{fadul2009collective}, reflecting concerns in our study. Switching to DOK to inform the AI for question generation helps alleviate this to a degree to meet our exploratory goals, but it is still possible that simply describing each level is not enough instruction for the AI to create an appropriate question. Future work could explore how educators design problems to derive a step-by-step framework for AI-driven question generation.

 We also acknowledge the constraints of our data input and evaluation methods. We used a single reference content as context for the AI in each study, and a single human evaluator. Given the exploratory nature of this study, though, it was important to limit the parameters to best explain variations in the quality. Moving forward, researchers can test and evaluate the output more systematically, given that our findings have highlighted the capabilities (and limitations) of current genAI. As such, our work serves as an important preliminary step in advancing question generation through AI for advanced math.

 \subsubsection{Acknowledgements}This work was supported by an exploratory Interdisciplinary Research Development award from McCAIS at McGill University.

\bibliographystyle{splncs04}
\bibliography{references}
\end{document}